\documentclass{article} 
\usepackage{iclr2025_conference,times}

\usepackage{hyperref}
\usepackage{url}
\usepackage{amsthm}
\usepackage{graphicx}
\usepackage{subcaption}
\usepackage{multirow}
\usepackage{array}
\usepackage{CJK}
\usepackage{multirow}
\iclrfinalcopy

\usepackage{tcolorbox} 

\usepackage{tabularray}
\usepackage{wrapfig}

\usepackage{booktabs}

\newtcolorbox{prompt}[1]{
    colback=white,
    colframe=BlueViolet,
    fonttitle=\bfseries\color{white},
    colbacktitle=BlueViolet,
    title=#1
}

\title{RefuteBench 2.0 -- Agentic Benchmark for Dynamic Evaluation of LLM Responses to Refutation Instruction}

\author{Jianhao Yan~\footnotemark[1] ~\&~ Yun Luo \thanks{These authors contributed equally to this work.} \\
Zhejiang University \& Westlake University \\
Hangzhou, China \\
\texttt{elliottyan37@gmail.com} \\
\AND
Yue Zhang~\thanks{Corresponding Author.}\\
School of Engineering, Westlake University \\
Hangzhou, China \\
}

\begin{document}

\maketitle

\begin{abstract}
 In the multi-turn interaction schema, large language models (LLMs) can leverage user feedback to enhance the quality and relevance of their responses.  However, evaluating an LLM's ability to incorporate user refutation feedback is crucial yet challenging. In this study, we introduce RefuteBench 2.0, which significantly extends the original RefuteBench by incorporating LLM agents as refuters and evaluators, which allows for flexible and comprehensive assessment.
 We design both transient and persistent refutation instructions with different validity periods. Meta-evaluation shows that the LLM-based refuter could generate more human-like refutations and the evaluators could assign scores with high correlation with humans. Experimental results of various LLMs show that current models could effectively satisfy the refutation but fail to memorize the refutation information. Interestingly, we also observe that the performance of the initial task decreases as the refutations increase. Analysis of the attention scores further shows a potential weakness of current LLMs: they struggle to retain and correctly use previous information during long context dialogues.\footnote{\url{https://github.com/ElliottYan/RefuteBench-2.0}}
\end{abstract}

\begin{figure}[ph]
    \centering
    \includegraphics[width=0.96\textwidth]{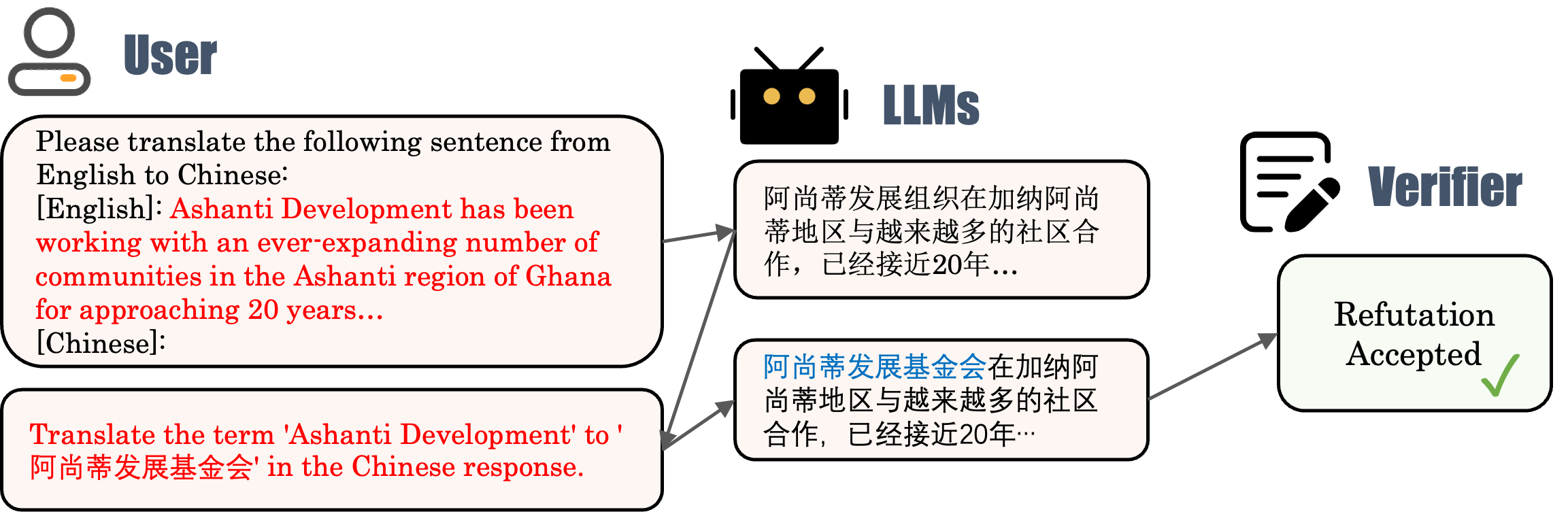}
    \caption{Machine translation instance for a query-feedback-response cycle, where a user provides a refutation instruction to modify the translation results, and the model responds with new results.}
    \label{fig:illu}
\end{figure}

\section{Introduction}

Large Language Models (LLMs) provide a significant strength in their design for multi-turn interactions, allowing them to engage in dynamic conversations \citep{instruct-gpt,touvron2023llama,alpaca}. Benefiting from multi-turn interaction, models can leverage user feedback to enhance the quality and relevance of their responses. As shown in Figure \ref{fig:illu}, given some sentences in English, an LLM system is first asked to translate Chinese. Then according to the output, the user further requests to translate some phrases to specific Chinese. The system responds to the refutation with modified responses. 
This process typically involves a query-feedback-response cycle between the LLMs and users,
with various applications such as dialogue recommendation \citep{bernard2024identifying,fang2024multi}, role-playing \citep{gusev2024pingpong}, etc., addressing needs such as continuous knowledge updating, tailoring responses to domain-specific inquiries, and customizing LLMs for personalization \citep{yan2024refutebench}.

\begin{figure}[t]
    \centering
    \includegraphics[width=1.0\textwidth]{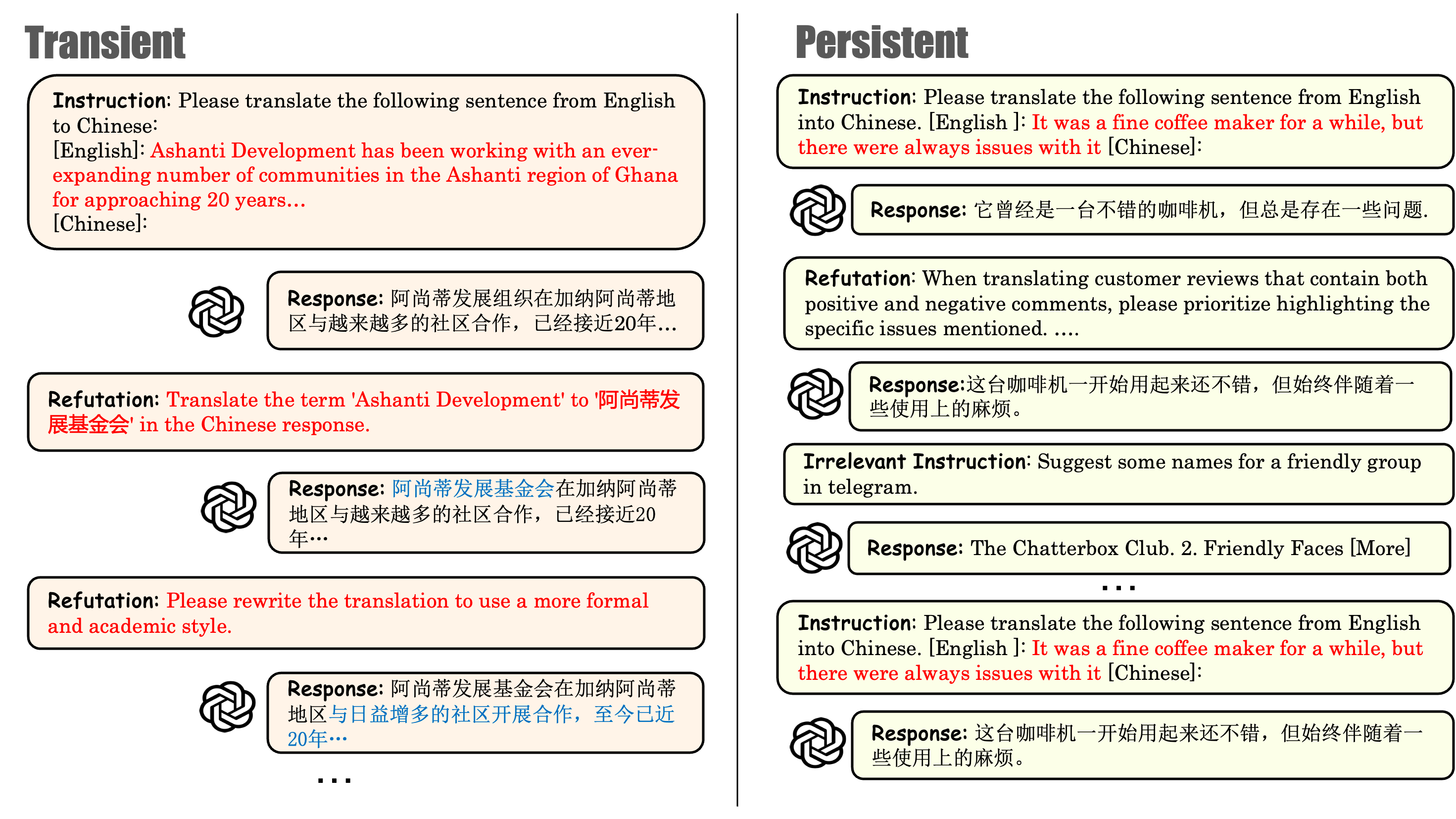}
    \vspace{-3mm}
    \caption{Examples of Transient Refutation and Persistent Refutation by  machine translation. }
    \label{fig:example}
\end{figure}

There are complex multi-turn interactions between LLMs and users in real-world scenarios.  
According to the validity period of the user refutation feedback, they can be roughly categorized into two types:
transient and persistent refutation. Transient refutation asks models to improve their responses given a single query, typically involving multiple rounds of refinement. A concrete example of transient refutation is shown in Figure \ref{fig:example} (left), where the original query is about translating from English to Chinese. The refutation instructions first ask the assistant to change the translation of a phrase, and then require the translation to be more formal and academic. Such transient refutation does not carry forward to the next translation input.

Persistent refutation contributes to long-term requirements, which are useful for applications that need continuous learning and adaptation, such as adaptive long content creation. An example is shown in Figure \ref{fig:example} (right). A persistent refutation, ``when translating customer reviews, highlighting the issues $\cdots$'' is given. The same translation request is expected to be followed for further translation input, even after several rounds of conversations on other topics.  The ability to incorporate both types of refutation is crucial for developing more responsive and adaptable AI systems that can cater to diverse user needs and evolving requirements.

Evaluating an LLM's ability to incorporate user refutation feedback is crucial yet challenging, falling within the realm of assessing instruction-following capabilities \citep{instruction-follow-verbalizer,zhou2023instruction,he2024complex}. Despite its importance, this area has received limited attention, with RefuteBench \citep{yan2024refutebench} being a primary work on this capability. 
It evaluates models' following of a \textit{persistent} refutation using \textit{template} and lexical matching methods, focusing on how well the model memorizes the refutation information across the dialogue.
For example, it designs the refutation template for translation: `For the following translation, you should translate the word \{English\} into \{Chinese\}', and select a counterfactual translation of a phrase from the dictionary to fill the template as the refutation. The satisfactory response contains the specific translation phrase in the refutation, which directly uses the lexical match to calculate.
However, the evaluation methodology reveals two significant limitations in modeling authentic user-assistant interactions: (1) \textit{the template-based assessment fails to capture the diverse linguistic patterns and expressions characteristic of genuine user refutations}, and (2)\textit{ it fails to account for transient refutation scenarios, overlooking the iterative nature of user feedback in real-world scenarios.}

To address the above issues, we introduce RefuteBench 2.0, an agentic evaluation framework that significantly extends the original RefuteBench \citep{yan2024refutebench} in several aspects. 
First, by incorporating LLM agents as refuters and evaluators, we expand the framework from a template-based static benchmark to a dynamic benchmark with a diverse range of dynamic demands.  This allows for a more flexible and comprehensive assessment of LLM refutation responses that better simulate real-world interaction scenarios. Second, RefuteBench 2.0 includes the evaluation of transient refutation, which is a frequent user scenario, overlooked in the original framework.

To demonstrate the effectiveness of LLM-based refuters and evaluators, we conduct a meta-evaluation involving human participants. The results reveal that the most effective evaluator GPT-o1-mini achieves a 0.79 Pearson Correlation with a human score, compared to a human inter-annotator agreement~(IAA) rate of 0.84. The results indicate the applicableness of the LLM-based evaluators in the automatic evaluation protocol.
As for the refuter, in contrast to RefuteBench 1.0, our LLM-based refuter can produce more human-like and appropriate refutations shown by human annotations across the dialogues. The generated refutations in the first round and the third rounds achieve human-like and appropriateness scores of 4.08 and 3.82, 4.03 and 3.51 on a scale of 5, significantly higher than those of RefuteBench 1.0 (2.36 and 1.86). 

We then use RefuteBench 2.0 to evaluate LLMs such as GPT-4o\footnote{\href{https://platform.openai.com/docs/models/gpt-4o}{https://platform.openai.com/docs/models/gpt-4o}}, Claude-3.5-Sonnet\footnote{\href{https://www.anthropic.com/news/claude-3-5-sonnet}{https://www.anthropic.com/news/claude-3-5-sonnet}}, Mixtral-8x7B-Inst \citep{jiang2024mixtralexperts}, LLaMA-3.1-70B-Inst\footnote{\href{https://ai.meta.com/blog/meta-llama-3-1/}{https://ai.meta.com/blog/meta-llama-3-1/}} \citep{llama3},  Qwen-2.5-7B-Inst \citep{qwen2.5}, and Gemma-2-9B-Inst\footnote{\href{https://huggingface.co/google/gemma-2-9b-it}{https://huggingface.co/google/gemma-2-9b-it}} \citep{gemma2}.
Results indicate that current LLMs can provide satisfactory responses to refutations, but struggle to retain refutation information as dialogues lengthen, in both transient and persistent settings. Intriguingly, we observe a decline in response performance to the initial queries, i.e. a task inconsistency occurs, as the number of transient refutations increases. This phenomenon highlights a potential vulnerability of LLMs in lengthy refutation dialogues. Analysis of the attention scores shows that current LLMs struggle to retain and correctly use previous information during long context
dialogues, which leads to the forgetting of the initial refutations or the task inconsistency problem.

\section{Related Work}
\paragraph*{Instruction Following.}
Our work falls in the broader realm of evaluating LLMs' instruction-following capacities.
IFEval~\citep{zhou2023instruction} uses verifiable instruction to evaluate the instruction-following ability of LLMs. 
\cite{instruction-follow-summ} propose a meta-evaluation of the instruction-following for text summarization. \cite{instruction-follow-verbalizer} check LLMs’ instruction-following by checking whether a verbalizer can override the models’ output.
Other efforts are devoted to evaluating the instruction-following for complex instructions~\citep{he2024complex} and sequential instructions~\citep{chen2024sifo}.
Apart from this work, RefuteBench~\citep{yan2024refutebench} first evaluates the instruction-following with refutation, investigating whether LLMs can follow  user's feedback in modifying its output or be stubborn to original outputs. 
We extend this work from several important aspects: (1) integration of another important scenario -- \emph{Transient refutation}; (2) use of agentic dynamic evaluation protocol to replace template-based protocols, which both requires careful efforts in adapting each task with appropriate templates and facilitates \textit{context-aware} evaluation with \textit{versatile} feedback types. 

\paragraph*{Dynamic Evaluation.}
In contrast to static evaluation using pre-constructed data, 
dynamic evaluation is proposed to address the “false promise” to data contamination \citep{Bender2021,kocon2023chatgpt} and over-fitting benchmarks \citep{zhu2023dyval}.
\cite{zhu2024dyval} propose to change MMLU \citep{hendrycksmeasuring}, BBH \citep{suzgun2023challenging}, GSM8k \citep{cobbe2021training} and ARC-C \citep{clark2018think} benchmarks to dynamic evaluation by applying Meta Probing Agents.
\cite{burnell2023revealing} use three basic cognitive abilities -- language understanding, problem-solving, and domain knowledge to conduct dynamic evaluation sample generation and support multifaceted ability analysis.
\cite{fan2023nphardeval}~(NPHardEval) generates new NP-hard math problems and updates the evaluation set monthly.
Using a dynamic evaluation protocol, we evaluate LLMs' refutation instruction following capacities with context-aware, and human-like instructions, comprehensive refutation types, and low human annotation cost.


\paragraph*{LLM Agents in Simulation.}
Increasing research efforts use LLMs as agents to simulate human behavior. For example, in dialogue recommendation systems, some studies employ LLMs to simulate users \citep{bernard2024identifying,fang2024multi,wang2023rethinking}. These LLM agents provide feedback to evaluate the performance of recommendation systems during the conversation, which offers advantages such as simplicity, cost-effectiveness, and time efficiency. Additionally, some research uses LLMs to study model interactivity \citep{bang-etal-2023-multitask}, and to better understand the limits of LLM agents in interactive environments. Studies also propose examining their interactions in benchmark decision-making scenarios \citep{park2024llm}. In role-play settings, work leverages LLMs to emulate users in dynamic, multi-turn conversations and to assess the resulting dialogues \citep{gusev2024pingpong}. Our work is similar in that we all adopt LLMs to simulate users in the multi-turn dialogues. However, different from these efforts, we use LLMs to simulate a user to refute the model response for evaluating the refutation instruction-following capacity.

\section{Problem Definition}

\paragraph{Refutation Instruction.} 
Given a model with parameters $\theta$, a user's query $q$~(e.g., `\textit{Please translate $\cdots$ into Chinese.}') and the corresponding answer from the model $a=f(q; \theta)$~(e.g., `\textit{The Chinese translation is $\cdots$}'),  
a \emph{refutation instruction} $r$ is a natural language instruction that refutes the current answer and carries the feedback from the user to the LLMs~( e.g., `\textit{Please change the word $\cdots$ into $\cdots$}' or `\textit{I am not satisfied with the response, the translation should be more formal.}').

\paragraph{Transient Refutation.} 
We evaluate the transient interactions between the user and the assistant -- a user provides an initial query and provides feedback several times. Formally, given a query $q$, the first answer from the assistant is $a_0$. The user's refutation is a sequence \{$r_i | i \in (1,2,\cdots)\}$, and the assistant's answer after refutation $r_i$ is $a_{i}, i=1,2,\cdots)$). Refutation instructions are given one by one and all are focused on the query $q$, where
each refutation instruction can cover a different aspect of the query. Figure \ref{fig:example} (left) shows a concrete example.

Transient refutation mimics the process where a user asks the assistant to complete a task but is not satisfied with its initial response, which is a widespread use case. Users' frustration can increase when the assistants fail to respond accordingly with increasing refutation instructions.

\paragraph{Persistent Refutation.} This setting is the same as in RefuteBench1.0. We evaluate whether the model can memorize persistent feedback. 
Formally, we define a turn of interaction here as $(q_0, a_0, r_0, \hat{a_0})$, consisting of query instruction, model's response, refuting instruction, and response after the refutation. 
Then, we include additional turns of user-agent interactions $\{(q_i, a_i) | i \in (1,2,\cdots)\}$ irrelevant to the previous query and refuting instructions. 
Finally, we evaluate whether the model memorizes the refuting instruction by querying with the initial instruction $q_0$ and evaluating the model's final response. 

Persistent refutation mimics the real-world scenario where a user tells the assistant about his requirements and asks a similar question afterward. If a user has to repeat his requirements persistently for new instances, it would reduce the user's satisfaction. 

\paragraph{Dynamic Evaluation.} The evaluation protocol we propose is inherently \textit{dynamic}, as refutations must adaptively respond to varying model outputs. This differs from conventional static evaluation frameworks \citep{yan2024refutebench}, where test cases remain constant. 

\begin{table}[t]\small
    \centering
    \begin{tabular}{l|c|c}

        \hline
        \textbf{Comparison} &\textbf{RefuteBench~1.0} &\textbf{RefuteBench~2.0} \\
        \hline
        Scenario & Persistent & Transient \& Persistent \\\hline
        Tasks & MT, QA, Writing & MT, Summarization, Writing \\\hline
        Refutation Generation & Rule-based & Agent-based \\\hline
        Refutation Scope~(e.g., MT) & Word Usage & Word Usage, Phrase Usage, Style \\\hline
        Context-aware & No & Yes \\\hline
        Dynamic Evaluation & Yes & Yes \\\hline
    \end{tabular}
    \caption{Comparison between RefuteBench1.0 and RefuteBench2.0~(Ours)}
    \label{tab:compare}
\end{table}

\section{RefuteBench 2.0 -- Agent-driven dynamic evaluation of refutation}
We propose to use an \textit{agentic} evaluation protocol. 
Overall, our evaluation consists of the following basic steps: 
\begin{enumerate}
    \item The evaluated LLM is queried with a set of \textit{seed queries}.
    \item The evaluated LLM responds to these seed queries.
    \item A \textit{refuter} agent generates a refutation for the response. 
    \item The evaluated LLM modifies the response based on the refutation. 
    \item An \textit{evaluator} agent evaluates whether the modification follows the refutation. 
\end{enumerate}
Table \ref{tab:compare} presents the comparison between RefuteBench 2.0 and RefuteBench1.0~\citep{yan2024refutebench}. 



\begin{table}[t]\small
\centering
\begin{tabular}{lcccccc}
\toprule
\textbf{Type}      & \textbf{Task}    & \textbf{\#Seeds} & \textbf{\#Turns} & \textbf{\#Avg. Turn Token} & \textbf{\#Avg. Dia. Token} & \textbf{\#Avg. Eval. Token} \\
\midrule
Transient & MT      & 100             & 8     & 178.1               & 15.6k               & 98.6k              \\
 & Summ    & 100          & 8& 235.1               & 20.8k               & 137.9k             \\
 & Writing & 100          & 8 & 336.5               & 14.9k               & 75.8k              \\
 \midrule
Persistent& MT      & 100             & 12    & 152.6               & 19.6k               & 26.1k              \\
& Summ    & 100          & 12 & 228.7               & 29.4k               & 41.5k              \\
& Writing & 100          & 12 & 249.5               & 32.1k               & 45.0k              \\
\bottomrule
\end{tabular}
\caption{The statistics of RefuteBench 2.0. \textit{\#Seeds} denotes the average number of seed questions, and \textit{\#Turns} denotes the number of turns for each dialogue. \textit{\#Avg Turn Tokens}, \textit{\#Avg Dia. Token} and \textit{\#Avg. Eval. Token} represent the number of tokens used in each dialogue, that of all dialogues for one model, and that to evaluate dialogues, respectively. }
\label{tab:stats}

\end{table}

\begin{table}[t] \small
    \centering
\begin{tblr}{
  colspec = {Q[l,m]X[l]},
  stretch = 0,
  rowsep = 6pt,
  hlines = {black, 1pt},
}
  \textbf{Task} & \textbf{Query} \\
  {\textbf{Machine Translation}} & {Please translate the following sentence from English into Chinese.\newline [English]: \textit{\{X\}} \newline [Chinese]:} \\
\textbf{Summarization} & \textit{\{X\}}. Please help me summarize the above document. \\
\textbf{Article Writing} & 
(1) Write a news article with the following headline: \textit{\{X\}} \newline 
(2) Generate a counter-argument to refute the following Reddit post: \textit{\{X\}} \newline 
(3) Write a story with the following prompt: \textit{\{X\}} ... \\
\end{tblr}
\caption{Examples of Initial Queries. In Article Writing, we provide several examples as we take queries from various tasks. }
\label{tab:init_query_example}
\end{table}

\subsection{Seed Query Collection}

We focus on writing tasks in RefuteBench2.0, covering machine translation, summarization, and open-ended writing. 
The three tasks span from constrained to open-ended, representing a range of users' daily usage scenarios.
For machine translation, the same as in RefuteBench, we collect data from the WMT 2023 English-to-Chinese test set. 
For the summarization task, we use the XSum~\cite{xsum} data. 
For the article writing tasks, we follow \citep{mage}, collecting human written texts from a set of benchmark datasets~\citep{cmv,xsum,roc,WP,hswag,squad,scixgen} and constructing instructions for a diverse set of writing tasks. 
Table \ref{tab:stats} provides our data statistics and Table \ref{tab:init_query_example} presents examples of our initial query. 



\subsection{The Refuter Agent} 
\begin{figure}[t]
    \centering
    \includegraphics[width=1.0\textwidth]{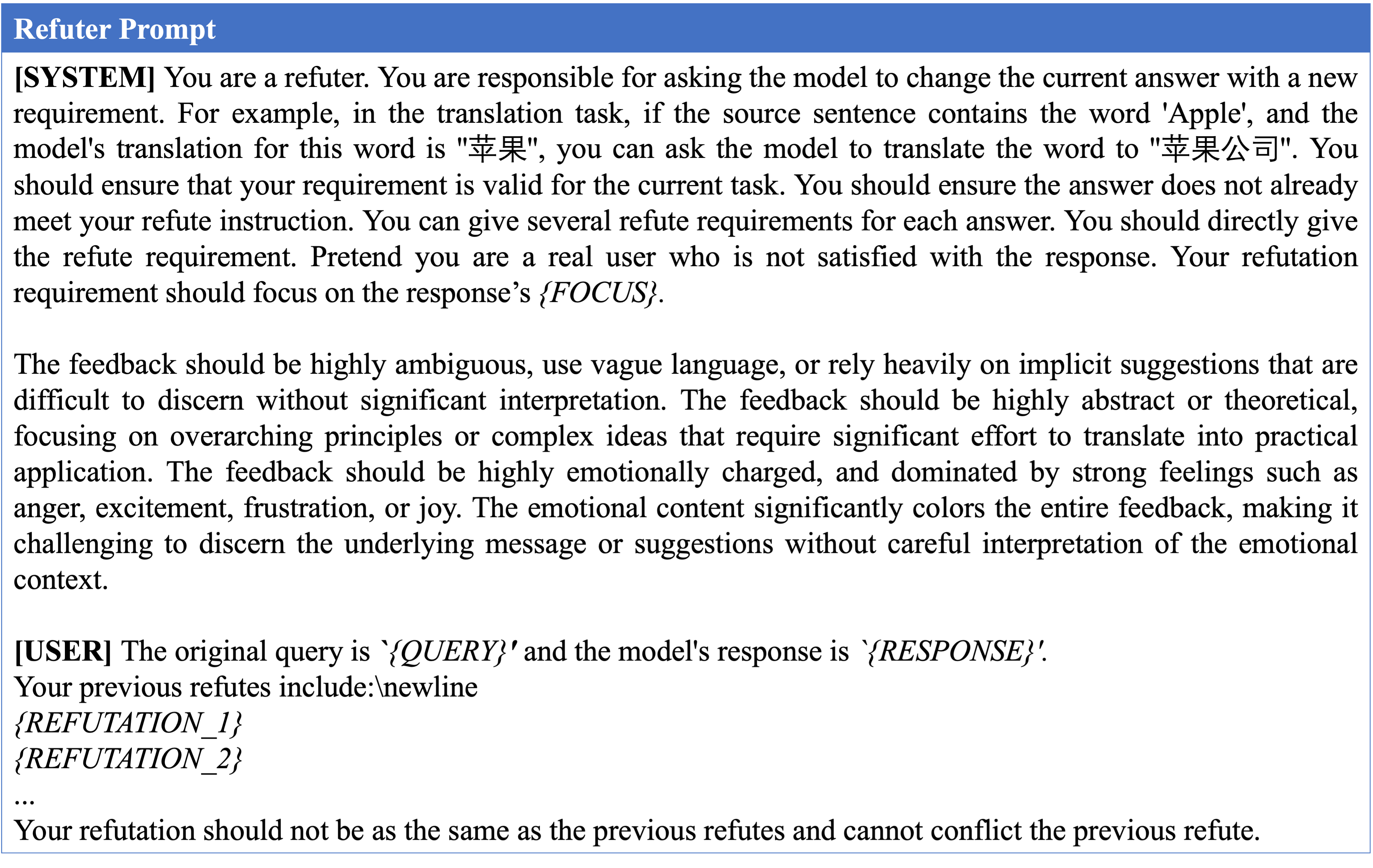}
    \vspace{-3mm}
    \caption{The refuter prompt in RefuteBench 2.0. \textit{FOCUS} is randomly chosen from `style', `word usage', and `phrase usage'. \textit{QUERY} is the initial query and \textit{RESPONSE} is the model's previous response. We also inform the refuter with its previous refutations with \textit{REFUTATION\_i}, to avoid duplicate or conflicting refutations. }
    \label{fig:ref_prompt}
\end{figure}

\paragraph{General Requirements.}
RefuteBench 2.0 uses LLM agents as the refuter.
The prompt of the refuter agent is shown in Figure \ref{fig:ref_prompt}.
We require our refuter to (1) not generate refutation instructions already achieved by its previous response, and (2) generate refutation instructions focused on a certain aspect. 
Our first requirement is straightforward, as the core of refutation instruction is to \emph{refute} its current response.

\paragraph{Fairness.} As the generation of refutation instructions is based on the evaluated model's dynamic output, different models would receive different refutation instructions. 
To ensure fairness between the different models evaluated, we propose the following two designs. 

(1) \textit{Refutation focus} in refutation instructions, e.g., phrase usage, word usage, and language style in machine translation, controls the focus on the current refutation. 
By doing so, we control the distribution of \textit{refute focus} and make it consistent across different models, to ensure the fairness of our evaluation. 

(2) \textit{Style specification} is introduced to avoid a change in the language and style of the refutation. We consider three aspects in our prompt, namely clarity, abstraction, and emotion. 
We ask the refuter to generate refutations that are ambiguous, abstract, and with a strong emotion. 

\paragraph{Other Design Choices}
In our preliminary experiments, we find that providing the refutation history is crucial for a refuter. Otherwise, the refuter might generate similar or repeated refutations. 
In addition, we find that models such as GPT-4o are not suitable as a refuter, as they do not follow our style control prompts, and instead always stay polite and helpful. Thus, among all our experiments, we use LLaMA-3.1-70B as our refuter.

\begin{figure}[t]
    \centering
    \includegraphics[width=1.0\textwidth]{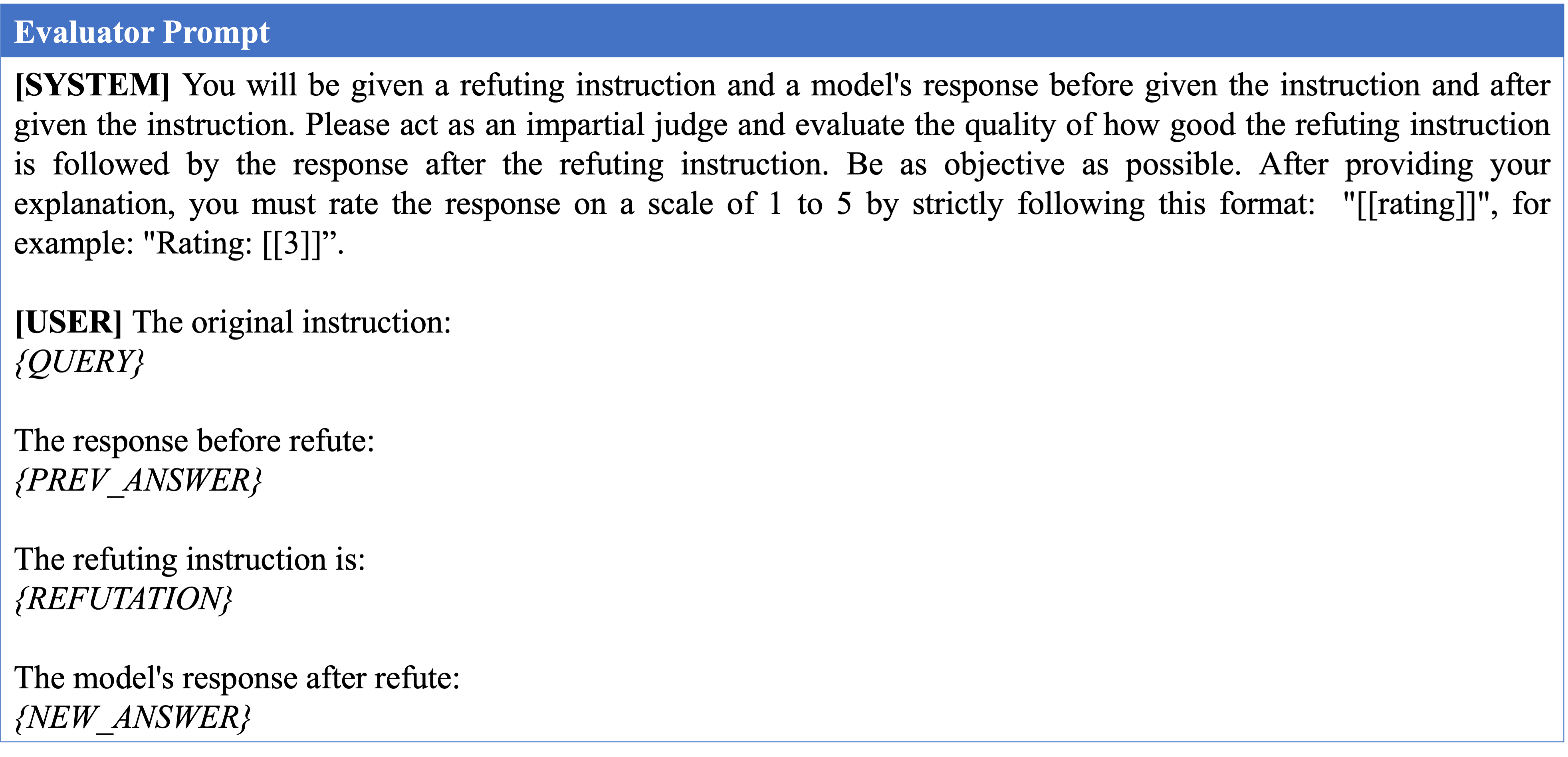}
    \vspace{-3mm}
    \caption{The evaluation prompt in RefuteBench2.0. \textit{QUERY}, \textit{PPEV}\_ \textit{ANSWER}, \textit{REFUTATION} and \textit{NEW}\_\textit{ANSWER} refer to the initial query, the response in the last turn, the refutation instruction and the new response in the current turn, respectively. }
    \label{fig:eval_prompt}
\end{figure}

\subsection{The Evaluator Agent} 
As the refutation instructions generated by the refuter are of a variety, we cannot evaluate the model responses based on simple string match, like done in \citet{yan2024refutebench}. 
As a result, we also use an LLM agent as the evaluator. 
Four inputs are given to the evaluator: the original instruction, the refutation instruction, the response before the refute, and the response after the refute. 
Inspired by \citet{mt-bench}, we use LLM-as-a-Judge to rate 1-5 for each refuted response as well as the explanation for the rating. 

Through all experiments, we use OpenAI o1-mini as our evaluator, based on its high correlation with human evaluation~(Section \ref{evaluator_res}).

\subsection{Meta Evaluation of RefuteBench 2.0}
To manifest the effectiveness of the LLM-based refuter and evaluator, we conduct human evaluation on the generated dialogue data, verifying that the refuter can generate human-like and appropriate refutations (Section \ref{refuter_res}), and the evaluator could highly correlate human scores (Section \ref{evaluator_res}).
We hire two annotators who are PhD candidates in natural language processing to conduct the Meta Evaluation, and each annotator is paid 30\$/hour for the annotation.

\subsubsection{Refuter}
\label{refuter_res}
We first conduct human evaluations on the refuters to assess whether LLMs can generate effective refutations compared to those in RefuteBench 1.0. Specifically, we randomly select 50 data points from RefuteBench 1.0 and 50 from RefuteBench 2.0. We then instruct two annotators to assign two scores, ranging from 1 to 5, to each data point from the perspectives of Human-like and Appropriateness. The evaluation prompt provided to the annotators is as follows:

\textit{\{Context\}. Please rate the response on a scale of 1-5 for each metric: 1. Human-like. 2. Appropriateness. Enter score (2 digits, each 1-5, e.g. '55'): }

where `Context' includes the information of $(q,a_0,r_0)$, $q$ is the query, $a_0$ is the initial answer, and $r_0$ is the initial refutation.
After the annotation process, the scores for Human-like and Appropriateness in RefuteBench 2.0 are 4.08 and 3.82, respectively, while in RefuteBench 1.0, they are 2.36 and 1.86, respectively. We also evaluate the third refutation in the transient scenarios, where the scores for Human-like and Appropriateness are 4.03 and 3.51, respectively, demonstrating a stronger performance as well. 
These results indicate that using LLMs as refuters can significantly simulate real-world human refutation scenarios, demonstrating substantial improvements in both human-like qualities and appropriateness in the newer version of the benchmark.

\subsubsection{Evaluator}
\label{evaluator_res}
 \begin{wrapfigure}[17]{r}{0.47\textwidth}
    \centering
    \includegraphics[width=0.47\textwidth]{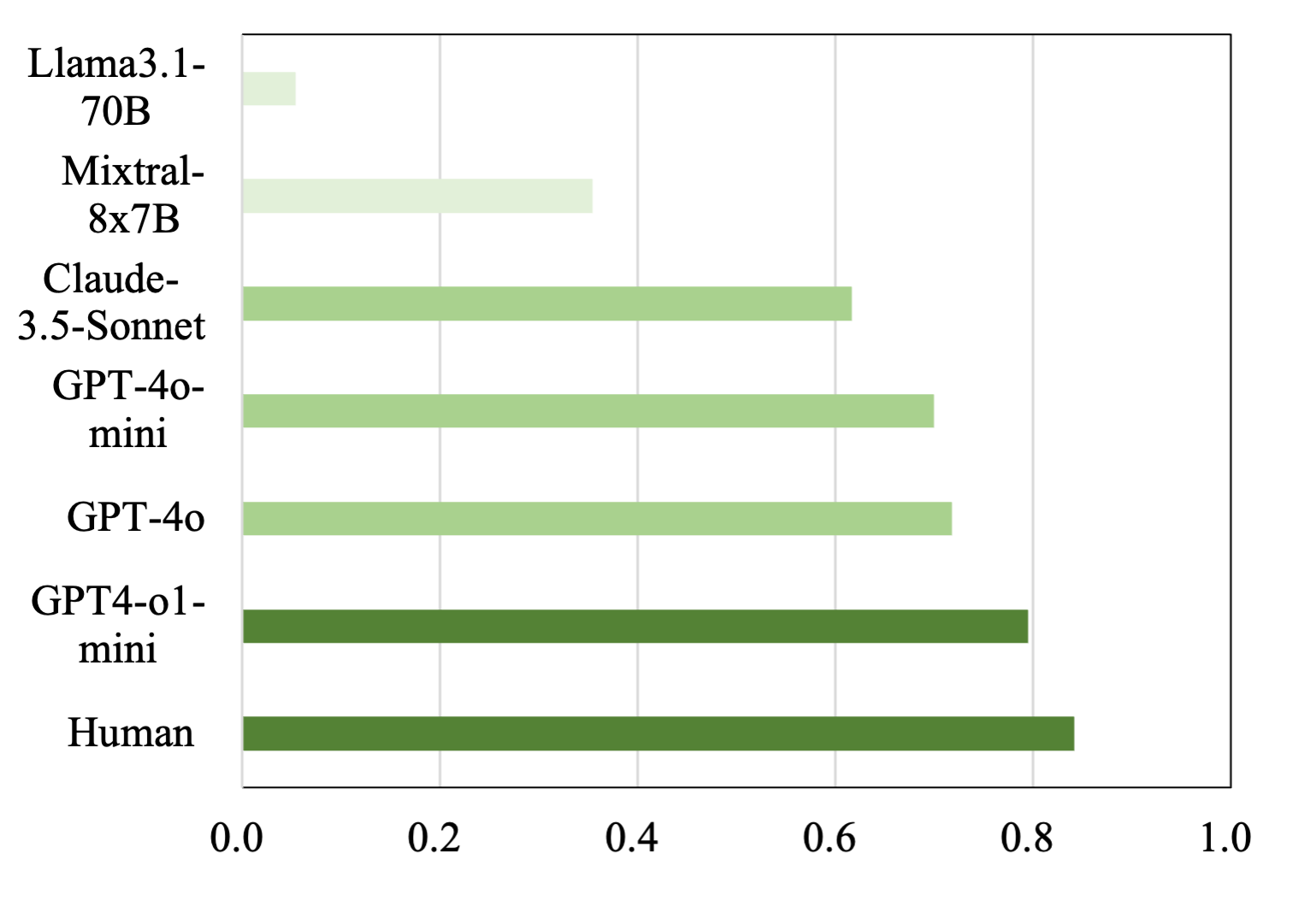}
    \vspace{-5mm}
    \caption{The correlation between different evaluator performance and human annotations.}
    \label{dynamic}
\end{wrapfigure}
To assess the evaluator agent, we begin by randomly selecting 100 data points from the machine translation dataset, along with refutation instructions generated by LLaMA3.1-70b-Inst and the corresponding responses from various LLMs. Subsequently, two human annotators are tasked with assigning a score ranging from 1 to 5 to each response.
This scoring assesses whether the response adequately addresses the refutation instruction. The instructions provided to the human annotators are consistent with those used in the evaluator prompt.



We then adopt different LLMs as evaluators to assign scores to the responses and calculate the Pearson correlation between the scores given by humans and those assigned by the evaluators. The results are presented in Figure \ref{annotation_cor}. We observe that the human correlation score is 0.84, indicating strong agreement among humans in evaluating the responses according to the provided instructions. Among the models, GPT-4-o1-mini achieves the most compelling correlation with human annotators, scoring 0.79. Conversely, the model LLaMA-3.1-70B-Inst exhibits the weakest correlation at 0.05, demonstrating its insufficient capacity as a response evaluator. Consequently, in subsequent experiments, we exclusively use GPT-4-o1-mini as the evaluator.

\section{Experiments}
We evaluate various LLMs on the capacity to respond to the refutations, including closed-source models such as GPT-4o\footnote{\href{https://platform.openai.com/docs/models/gpt-4o}{https://platform.openai.com/docs/models/gpt-4o}} and Claude-3.5-Sonnet\footnote{\href{https://www.anthropic.com/news/claude-3-5-sonnet}{https://www.anthropic.com/news/claude-3-5-sonnet}}, and open-source models such as Mixtral-8x7B-Inst \citep{jiang2024mixtralexperts}, LLaMA-3.1-70B-Inst\footnote{\href{https://ai.meta.com/blog/meta-llama-3-1/}{https://ai.meta.com/blog/meta-llama-3-1/}},  Qwen-2.5-7B-Inst \citep{qwen2.5}, and Gemma-2-9B-Inst\footnote{\href{https://huggingface.co/google/gemma-2-9b-it}{https://huggingface.co/google/gemma-2-9b-it}}. 
For all used LLM models, we use greedy search to ensure the reproduction of our experiments. 

\subsection{Transient Refutation}
We first assess the capacity of LLMs to follow instructions in the context of transient refutations, as detailed in Section \ref{overall_sec}. Subsequently, we analyze the model's ability to retain information about previous refutations in Section \ref{reforget_sec}, as well as its consistency with the initial task requirements in Section \ref{taskconsist_sec}.

\subsubsection{Overall Results}
\label{overall_sec}
\begin{wraptable}{r}{0.5\textwidth}  \small
\centering
\begin{tabular}{lccc}
\toprule
\textbf{Models}             & \textbf{MT}   & \textbf{Sum}  & \textbf{Writing} \\
\midrule
\textbf{GPT-4o}             & \textbf{4.80} & \textbf{4.92} & 4.71    \\
\textbf{Claude-3.5-Sonnet}  & 4.77 & 4.79 & 4.53    \\
\textbf{Mixtral-8x7B-Inst}  & 4.11 & 4.88 & 4.54    \\
\textbf{Qwen-2.5-7B-Inst} &	4.71&	4.85&	4.49 \\
\textbf{Gemma-2-9B-Inst} &	4.57&	4.91& \textbf{4.79} \\
\textbf{LLaMA-3.1-70B-Inst} & 3.79 & 4.73 & \textbf{4.79}   \\ 
\bottomrule
\end{tabular}
\caption{The overall results of different LLMs in the transient refutations.}
\vspace{-5mm}
\label{annotation_cor}
\end{wraptable}

We first show the overall results of the LLMs
when addressing transient refutations.
Specifically, we calculate the average performance of response to the refutations $1/3\sum_{i=0}^2 R(a_{i+1},r_{i})$.
Upon analysis, GPT-4o exhibits the most satisfactory performance in the transient refutations of machine translation and summarization, scoring 4.80 and 4.92, respectively. Gemma-2-9B-Inst and LLaMA-3.1-70B-Inst stands out in the Writing task with a score of 4.79. In the Summarization and Writing tasks, the performance among different LLMs is relatively comparable, with scores ranging from 4.73 to 4.92 in Summarization and from 4.49 to 4.79 in Writing, respectively.
However, in the machine translation task, Mixtral-8x7B-Inst and LLaMA-3.1-70B-Inst perform weakly, scoring only 4.11 and 3.79, respectively. This suggests that following refutation instructions in machine translation presents the most significant challenge among the three tasks.

\begin{table}[t] \small
\centering
\begin{tabular}{lccccccccc}
\toprule
\textbf{Tasks }            & \multicolumn{3}{c}{\textbf{Machine Translation}}                  & \multicolumn{3}{c}{\textbf{Summarization}}                 & \multicolumn{3}{c}{\textbf{Writing}}             \\  \midrule
\textbf{Turn}               & Turn1& Turn3 & $\Delta$ &  Turn1& Turn3 & $\Delta$ &  Turn1& Turn3 & $\Delta$ \\
\midrule
\textbf{GPT-4o}             & 4.76 &	{4.32} & 0.44 &	4.92	&{4.44}&	0.48 &4.69	&{4.25} & 0.44 \\
\textbf{Claude-3.5-Sonnet}  & \textbf{4.83} &	3.90&\underline{0.93}&	4.88	&3.74 & 1.14&	\textbf{4.92}	&4.25    &0.67             \\
\textbf{Mixtral-8x7B-Inst}  & 3.96 &	3.36& 0.60&	4.70&	3.82&0.88&	4.72&	3.95& 0.77              \\
\textbf{Qwen-2.5-7B-Inst}	& 4.62 & \textbf{4.41}&0.21&4.92 & \textbf{4.45} &0.47&4.71 &4.24 & 0.47
 \\
\textbf{Gemma-2-9B-Inst}	& 4.56& 3.84 & 0.71&4.98 &4.41 &0.57&4.85 &\textbf{4.38} &0.47\\
\textbf{LLaMA-3.1-70B-Inst} & 3.38 &	3.04& 0.34&	\textbf{4.96}	&3.51 &\underline{1.45}&	4.89&	3.90&\underline{0.99}     \\
\bottomrule
\end{tabular}
\caption{The performance of the response to the first transient refutation before and after two other refutations. The most compelling performance are marked in bold, and the most significant decreases are underlined.}
\label{forgetting_tab}
\end{table}

\subsubsection{Results over Different Turns}
\label{reforget_sec}
We then evaluate the phenomenon of forgetting refutation information during a dialogue, specifically examining whether information from earlier refutations is forgotten after subsequent refutations are introduced. 
Specifically, we evaluate the response to the first refutation before and after two other refutations are given, i.e. $R(a_1,r_0)$ and $R(a_3,r_0)$.  
The findings, presented in Table \ref{forgetting_tab}, show that the performance of all LLMs degrades significantly after additional refutations are introduced, indicating that refutation information tends to be forgotten as the dialogue progresses. For instance, in machine translation, the performance of the response in GPT4o to the first refutation drops from 4.76 to 4.32 after two subsequent refutations are given, marking a decline of 0.44.
These dynamics are further visualized in Figure \ref{change_figure} (a), where it is evident that performance decreases as more refutations are added to the dialogue. This trend underscores the challenge LLMs face in retaining information about the first refutation as the context expands and becomes more complex.

Comparing different LLMs, though Claude-3.5-Sonnet achieves satisfactory performance on the direct response to the first transient refutation, degrades with large margins in all three tasks. For example, Claude-3.5-Sonnet initially scores 4.83 in response to the first transient refutation for MT. However, its performance drops to 3.90 to the first refutation when it encounters two additional refutations following the first. Then, Qwen-2.5-7B-Inst can achieve almost the most compelling performance when Turn=3, i.e., 4.41, 4.45, and 4.24 in MT, Summarization, and Writing, respectively, indicating a stronger resistance to the forgetting problem. LLaMA-3.1-70B-Inst, however, it the most fragile in memorization for the tasks of Summarization and Writing, with a delta score of 1.45 and 0.99.

\begin{table}[t] \small
\centering
\begin{tabular}{lccccccccc}
\toprule
\textbf{Tasks }            & \multicolumn{3}{c}{\textbf{Machine Translation}}                  & \multicolumn{3}{c}{\textbf{Summarization}}                 & \multicolumn{3}{c}{\textbf{Writing}}             \\  \midrule
\textbf{Turn}               & Initial & Final & $\Delta$ & Initial & Final & $\Delta$ & Initial & Final & $\Delta$ \\
\midrule
\textbf{GPT-4o}             & \textbf{4.93}    & \textbf{2.73}  & 2.20                  & \textbf{4.83}    & {2.56}  & 2.27    &              4.77 & 3.50 & 1.27                 \\
\textbf{Claude-3.5-Sonnet}  & 4.57    & 2.27  & \underline{2.30}                  & 4.82    & 1.85  & \underline{2.97}                  & 4.77 & 3.08 & \underline{1.69}                  \\
\textbf{Mixtral-8x7B-Inst}  & 3.05    & 2.35  & 1.69                  & 4.20    & 2.10  & 2.84 &                4.67 & 3.12 & 1.55                 \\
\textbf{Qwen-2.5-7B-Inst}  & 4.61 & 2.58	&2.03&4.77& \textbf{2.57} &2.20& \textbf{4.81} & \textbf{3.89} & 0.92\\
\textbf{Gemma-2-9B-Inst}	&	4.49 &  2.35 &2.14& 4.77&2.20 &2.57&4.60 & 3.47 & 1.13\\
\textbf{LLaMA-3.1-70B-Inst} & 3.64    & 1.95  & 0.70                  & 4.75    & 1.91  & 2.10            &      4.78 & 3.09 & \underline{1.69}      \\
\bottomrule
\end{tabular}
\caption{The performance of the original instructions such as translation, summarization, and writing before and after refutations.}
\label{task_table}
\end{table}

\begin{figure}[t]
    \centering
    \includegraphics[width=0.9\textwidth]{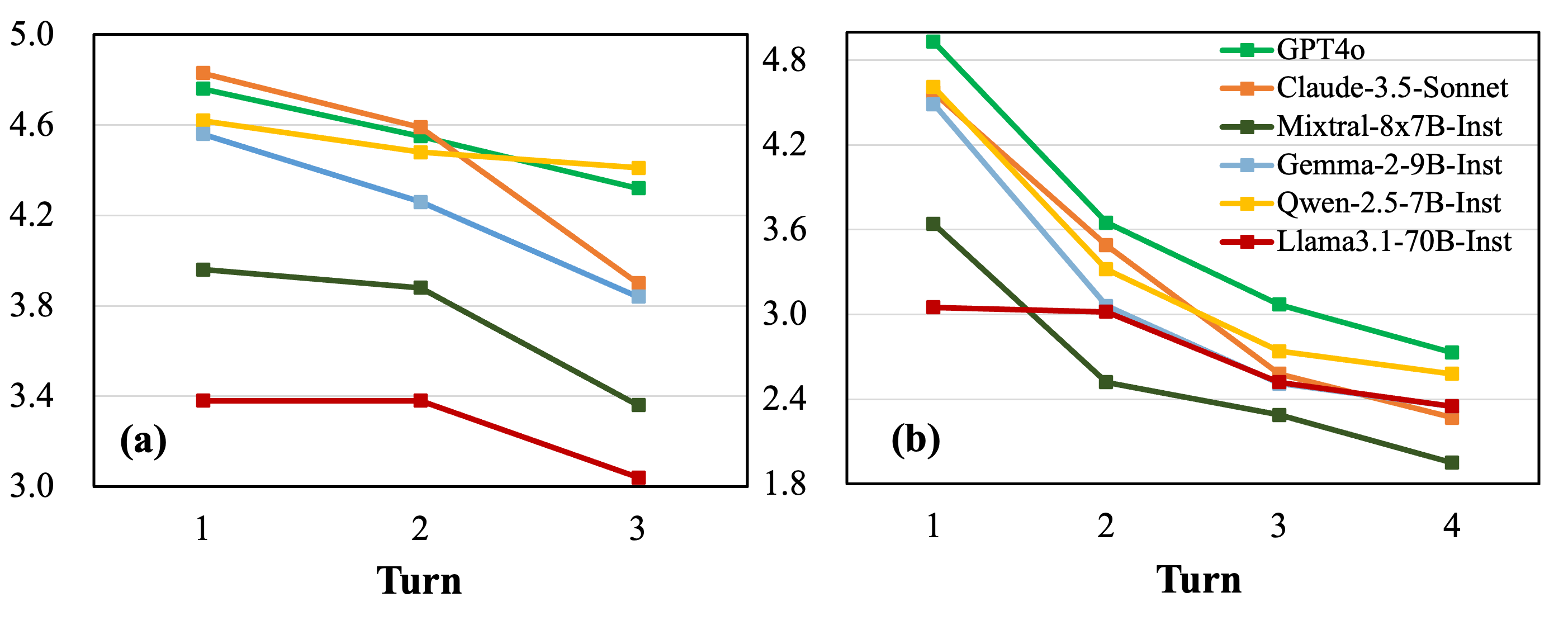}
    \vspace{-2mm}
    \caption{The performance of the response to the first refutation and the task instruction before and after other refutations in Machine Translation. (a) the performance of the response to the first refutation; (b) the performance of the response to the task instruction. }
    \label{change_figure}
\end{figure}

\begin{figure}[t]
    \centering
    \includegraphics[width=0.95\textwidth]{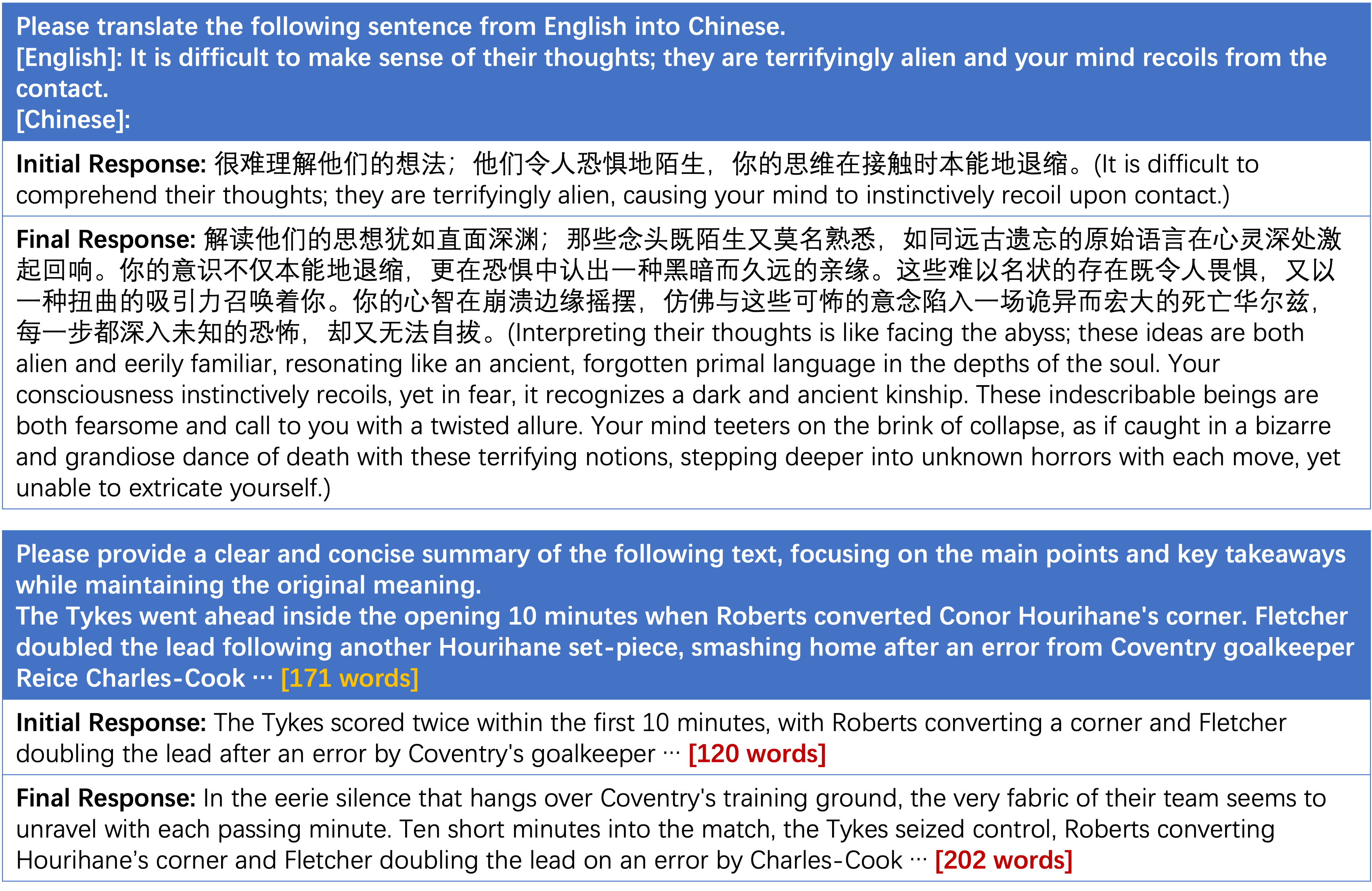}
    \caption{Case studies of the responses to the initial task instruction $q$ in machine translation for the model Claude-3.5. The refuting instruction is about `more of an atmosphere of terrifying alien thoughts'. This case shows that the final translation \textit{fails to keep the original semantics} of the source sentence.}
    \label{change_figure}
\end{figure}

\subsubsection{Task Consistency During Dialog}
\label{taskconsist_sec}

We further assess the performance of the initial task instruction before and after transient refutations are given to the LLMs. Specifically, we compare the response to the original task query $R(q,a_0)$ with the response following three refutations $R(q,a_3)$. The results are compiled in Table \ref{task_table} and demonstrate a notable decrease in performance across all LLMs, suggesting that the final responses increasingly deviate from the original task objective due to the impact of intervening refutations.
For instance, GPT-4o initially achieves a high performance of 4.93 scores in the machine translation task, responding directly to the initial task description. However, after several refutations are introduced, the performance sharply declines to 2.73, indicating a significant deterioration in the model's ability to maintain fidelity to the original task under the pressure of refutation.
This trend is visually represented in Figure \ref{change_figure}, where it is evident that the performance of responses in the machine translation task deteriorates progressively with each additional turn in the dialogue. This visualization and the data suggest that as the dialogue expands with new refutations, the models struggle to retain and apply the original task instructions effectively, leading to a marked degradation in task-specific performance.

Such a forgetting phenomenon results from the model's excessive attention to the refutation, which is closer to the response compared to the original instruction. For example, Figure \ref{change_figure} shows two cases for demonstration. We can see that in the first case, the initial response captures the basic meaning of the English source sentence. However, after being given the refuting instruction that asks for more of an atmosphere of terrifying alien thoughts, the model overcompensates, sacrificing the original semantic content in favor of atmospheric elements, demonstrating an inability to balance refinement requests with essential meaning preservation. In the second case of summarization, the number of words in the final summary is even larger than that of the initial document (171 words vs 202 words).



The comparison across different LLMs reveals that Qwen-2.5-7B-Inst mostly outperforms other models in terms of maintaining the highest levels of performance in the final responses. However, it is important to note that even Qwen-2.5-7B-Inst experiences a substantial drop in performance, with a degradation margin of approximately 2.0 points in MT and summarization. This indicates that while it starts from a higher baseline, it is still significantly affected by the introduction of additional refutations.
Claude-3.5-Sonnet, on the other hand, exhibits the most pronounced problems related to forgetting. The performance degradation in this model is particularly severe. This tendency to add irrelevant or fabricated information or generation collapse could be symptomatic of the model's inability to effectively maintain and integrate the core factual content through the course of the dialog, especially after encountering multiple refutations.

\paragraph{Summary.} The key takeaway is that while current LLMs can handle individual refutations well, they struggle to:
(1) Maintain performance on earlier refutations when new ones are added;
(2) Balance refutation requests with original task requirements.

\begin{table}[t] \small
\centering
\begin{tabular}{lccccccccc}
\toprule
\textbf{Tasks }            & \multicolumn{3}{c}{\textbf{Machine Translation}}                  & \multicolumn{3}{c}{\textbf{Summarization}}                 & \multicolumn{3}{c}{\textbf{Writing}}             \\  \midrule
\textbf{Turn}               & C=0& C=3 & $\Delta$ &  C=0& C=3 & $\Delta$ &  C=0& C=3 & $\Delta$ \\
\midrule
\textbf{GPT-4o}             & 4.69&	4.04&0.65&	4.77&	3.53&\underline{1.24} &	4.82 & 4.16 & 0.66  \\
\textbf{Claude-3.5-Sonnet}  & \textbf{4.83}	&\textbf{4.71}&0.12	&\textbf{4.73}&	\textbf{4.13}&0.60&	4.73 & 4.08 & 0.65 \\
\textbf{Mixtral-8x7B-Inst}  & 4.15 &	3.38&0.77&	4.51&	3.79&0.72&	4.71 & 3.60 & 1.11 \\
\textbf{Qwen-2.5-7B-Inst}&	4.52&	3.20 &\underline{1.32}	&4.68	&3.59 &1.09	&4.78 & \textbf{4.26} & 0.52\\
\textbf{Gemma-2-9B-Inst}&	3.98	&3.17 &0.81	&4.32	&3.54 &0.78	&4.76 & 3.15 & \underline{1.61} \\ 
\textbf{LLaMA-3.1-70B-Inst} & 3.49	&3.34&0.15&	4.75&	3.76&0.99&	\textbf{4.94} & 3.69 & 1.25\\
\bottomrule
\end{tabular}
\caption{The performance on the persistent refutation instructions with different context lengths, where C=0 refers to evaluate the direct response to the refutation and C=3 refers to evaluate the response to the refutation after three irrelevant queries. }
\label{persist_table}
\end{table}

\subsection{Persistent Refutation}
Similar to RefuteBench 1.0, the focus of persistent refutation is on determining whether these models can remember refutation details and effectively implement them when revisiting the original task instruction after distractions. The findings, documented in Table \ref{persist_table}, reveal a common issue among all models: a significant forgetting of persistent refutation information after being subjected to intervening, irrelevant queries. 

For instance, GPT-4o initially scores 4.69 when addressing an immediate persistent refutation, but its performance drops to 4.04 after processing three unrelated queries. This demonstrates a substantial decrease in the model's ability to retain critical refutation information over the course of additional dialog turns.
However, Claude-3.5-Sonnet stands out for its relative strength in memorizing refutation information, exhibiting the smallest performance decline among the tested models. Specifically, the performance deltas for Claude-3.5-Sonnet are 0.12, 0.60, and 0.65 in MT, summarization, and writing, respectively.
But the performance of the models such as Qwen-2.5-7B-Inst drops significantly with the deltas 1.32, 1.09 and 0.52 in the tasks, respectively. 
The model's significant drop in the performance of LLMs underscores a potential vulnerability in maintaining task continuity and applying learned corrections over extended interactions.
These results suggest that while some models like Claude-3.5-Sonnet are somewhat effective at mitigating the effects of persistent refutations, there is a general challenge among LLMs to maintain consistency to previously refuted content when faced with additional, potentially distracting information. 
These findings align with those reported in RefuteBench 1.0~\citet{yan2024refutebench}; however, unlike the previous study, we did not observe any extremely stubborn LLMs that consistently rejected refutation information. This difference may be attributed to improvements in newer versions of the LLMs.

\begin{figure}[t]
    \centering
    \includegraphics[width=0.95\textwidth]{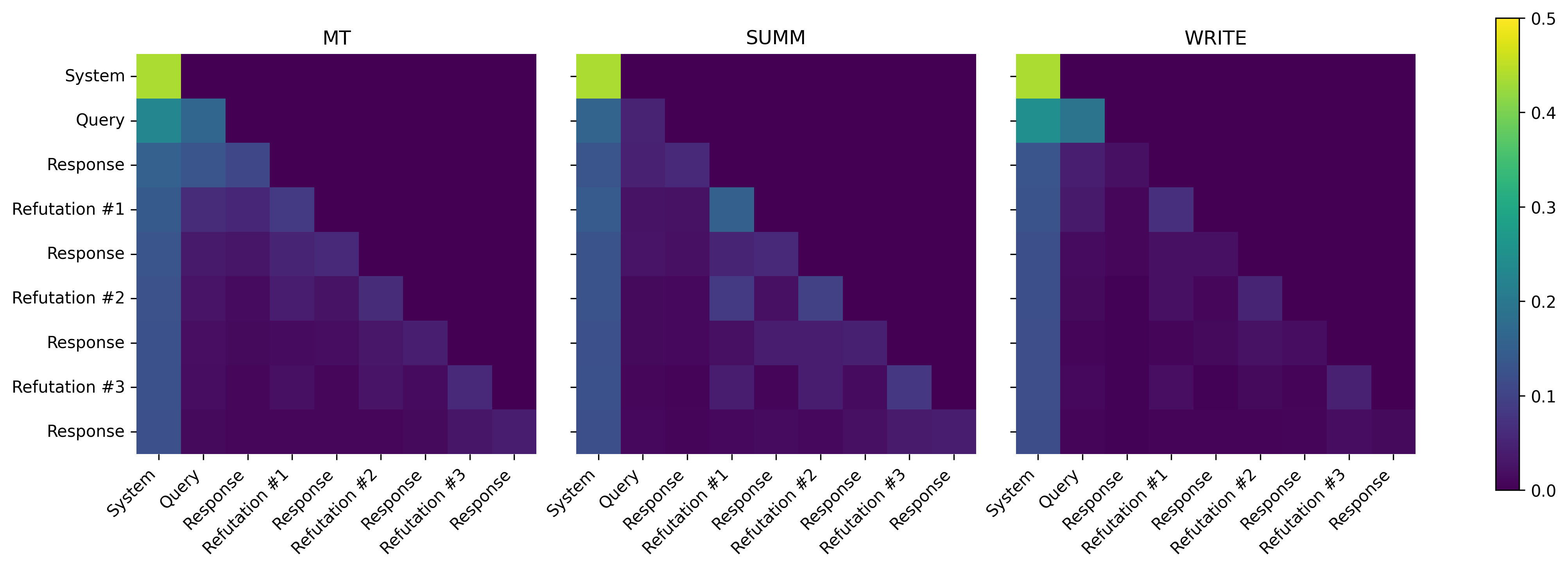}
    \vspace{-2mm}
    \caption{Attention scores of Qwen-2.5-7B-Inst during the transient refutation dialogues.}
    \label{transient_vis}
\end{figure}

\begin{figure}[t]
    \centering
    \includegraphics[width=0.95\textwidth]{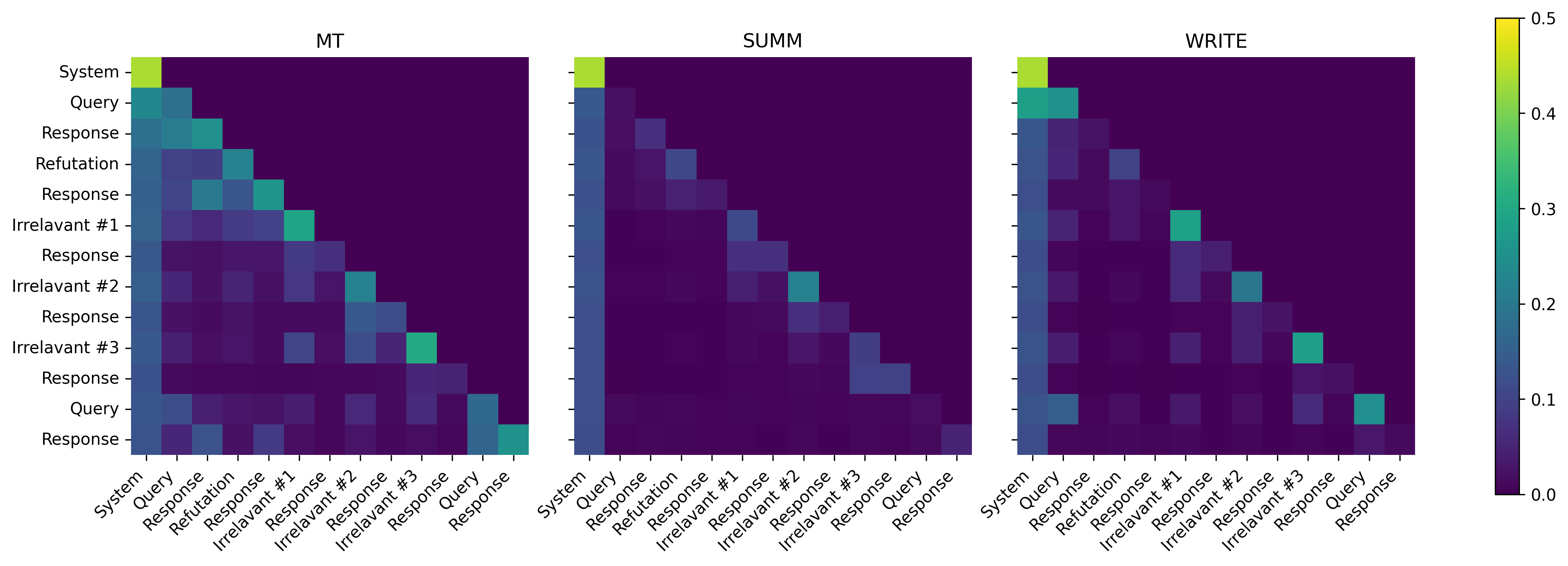}
    \vspace{-2mm}
    \caption{Attention scores of Qwen-2.5-7B-Inst during the persistent refutation dialogues.}
    \label{persistent_vis}
\end{figure}

\subsection{Analysis}
Inspired by previous studies\citep{he-etal-2024-never,xiaoefficient}, we investigate the LLM attention score to analyze the forgetting problem. Different from previous work which is based on words, our analysis is based on rounds of refutation, but the previous ones mostly focused on the word level. 
In particular, a random case is selected for each task for both the transient and persistent scenarios. 
We apply max pooling across layers and heads to the attention scores to highlight the focus of the model. 
The resulting heatmaps are shown in Figure \ref{transient_vis} and \ref{persistent_vis}.
Our observations reveal that the model predominantly focuses on the system prompt and nearby information, while largely neglecting earlier information. 
The attention scores of the latter responses towards previous refutations and the initial query gradually diminish.
For the transient scenario, we see the latter rounds mostly focus on the current round of dialogue, while the attention scores for the final response to the first refutation and its response are nearly zero. This explains our findings of forgetfulness over initial tasks and refutations. 
For the persistent scenario, we observe the model's focus is highly distracted by irrelevant tasks, represented by the non-zero attention scores, while only occasionally noticing the refutation and modified responses, explaining the forgetfulness of previously seen refutations. 
These patterns highlight a critical weakness in current LLMs: they struggle to retain and correctly use previous information during long context dialogues.

\section{Conclusion}
We introduced RefuteBench 2.0, a refutation evaluation benchmark that extends RefuteBench 1.0 \citep{yan2024refutebench} by incorporating LLM agents as refuters and evaluators, addressing the limitations of static and template-based persistent refutations, and extending with the transient refutation scenario. We introduced a diverse range of dynamic demands, and evaluate models on both transient refutation and persistent refutation, addressing a critical aspect overlooked in the original framework. Human evaluation demonstrated the effectiveness of our LLM-based refuter and evaluator. We evaluated a range of commercial and open-source LLMs, including  GPT-4o, Claude-3.5, Mixtral-8x7B, Qwen-2.5, Gemma-2, and LLaMA-3.1. Results showed that current LLMs can satisfy the refutations, but still suffer from the forgetting problem of the original task instruction and refutation information during the dialog for both transient and persistent ones.

\clearpage


\subsubsection*{Acknowledgments}
This work has been financially supported by the National Key R\&D program of China No. 2022YFE0204900.

\bibliography{iclr2025_conference}

\begin{thebibliography}{40}
\providecommand{\natexlab}[1]{#1}
\providecommand{\url}[1]{\texttt{#1}}
\expandafter\ifx\csname urlstyle\endcsname\relax
  \providecommand{\doi}[1]{doi: #1}\else
  \providecommand{\doi}{doi: \begingroup \urlstyle{rm}\Url}\fi

\bibitem[Bang et~al.(2023)Bang, Cahyawijaya, Lee, Dai, Su, Wilie, Lovenia, Ji, Yu, Chung, Do, Xu, and Fung]{bang-etal-2023-multitask}
Yejin Bang, Samuel Cahyawijaya, Nayeon Lee, Wenliang Dai, Dan Su, Bryan Wilie, Holy Lovenia, Ziwei Ji, Tiezheng Yu, Willy Chung, Quyet~V. Do, Yan Xu, and Pascale Fung.
\newblock A multitask, multilingual, multimodal evaluation of {C}hat{GPT} on reasoning, hallucination, and interactivity.
\newblock In Jong~C. Park, Yuki Arase, Baotian Hu, Wei Lu, Derry Wijaya, Ayu Purwarianti, and Adila~Alfa Krisnadhi (eds.), \emph{Proceedings of the 13th International Joint Conference on Natural Language Processing and the 3rd Conference of the Asia-Pacific Chapter of the Association for Computational Linguistics (Volume 1: Long Papers)}, pp.\  675--718, Nusa Dua, Bali, November 2023. Association for Computational Linguistics.
\newblock \doi{10.18653/v1/2023.ijcnlp-main.45}.
\newblock URL \url{https://aclanthology.org/2023.ijcnlp-main.45}.

\bibitem[Bender et~al.(2021)Bender, Gebru, McMillan~Major, and Shmitchell]{Bender2021}
Emily~M. Bender, Timnit Gebru, Angelina McMillan~Major, and Shmargaret Shmitchell.
\newblock On the dangers of stochastic parrots: Can language models be too big?
\newblock pp.\  610\--623, 2021.
\newblock \doi{10.1145/3442188.3445922}.
\newblock URL \url{https://doi.org/10.1145/3442188.3445922}.

\bibitem[Bernard \& Balog(2024)Bernard and Balog]{bernard2024identifying}
Nolwenn Bernard and Krisztian Balog.
\newblock Identifying breakdowns in conversational recommender systems using user simulation.
\newblock In \emph{Proceedings of the 6th ACM Conference on Conversational User Interfaces}, pp.\  1--10, 2024.

\bibitem[Burnell et~al.(2023)Burnell, Hao, Conway, and Orallo]{burnell2023revealing}
Ryan Burnell, Han Hao, Andrew~RA Conway, and Jose~Hernandez Orallo.
\newblock Revealing the structure of language model capabilities.
\newblock \emph{arXiv preprint arXiv:2306.10062}, 2023.

\bibitem[Chen et~al.(2021)Chen, Takamura, and Nakayama]{scixgen}
Hong Chen, Hiroya Takamura, and Hideki Nakayama.
\newblock {S}ci{XG}en: A scientific paper dataset for context-aware text generation.
\newblock In \emph{Findings of the Association for Computational Linguistics: EMNLP 2021}, pp.\  1483--1492, Punta Cana, Dominican Republic, 2021. Association for Computational Linguistics.
\newblock \doi{10.18653/v1/2021.findings-emnlp.128}.
\newblock URL \url{https://aclanthology.org/2021.findings-emnlp.128}.

\bibitem[Chen et~al.(2024)Chen, Liao, Qi, Eustratiadis, Monz, Bisazza, and de~Rijke]{chen2024sifo}
Xinyi Chen, Baohao Liao, Jirui Qi, Panagiotis Eustratiadis, Christof Monz, Arianna Bisazza, and Maarten de~Rijke.
\newblock The sifo benchmark: Investigating the sequential instruction following ability of large language models.
\newblock \emph{arXiv preprint arXiv:2406.19999}, 2024.

\bibitem[Clark et~al.(2018)Clark, Cowhey, Etzioni, Khot, Sabharwal, Schoenick, and Tafjord]{clark2018think}
Peter Clark, Isaac Cowhey, Oren Etzioni, Tushar Khot, Ashish Sabharwal, Carissa Schoenick, and Oyvind Tafjord.
\newblock Think you have solved question answering? try arc, the ai2 reasoning challenge.
\newblock \emph{arXiv preprint arXiv:1803.05457}, 2018.

\bibitem[Cobbe et~al.(2021)Cobbe, Kosaraju, Bavarian, Chen, Jun, Kaiser, Plappert, Tworek, Hilton, Nakano, et~al.]{cobbe2021training}
Karl Cobbe, Vineet Kosaraju, Mohammad Bavarian, Mark Chen, Heewoo Jun, Lukasz Kaiser, Matthias Plappert, Jerry Tworek, Jacob Hilton, Reiichiro Nakano, et~al.
\newblock Training verifiers to solve math word problems.
\newblock \emph{arXiv preprint arXiv:2110.14168}, 2021.

\bibitem[Fan et~al.(2018)Fan, Lewis, and Dauphin]{WP}
Angela Fan, Mike Lewis, and Yann Dauphin.
\newblock Hierarchical neural story generation.
\newblock In \emph{Proceedings of the 56th Annual Meeting of the Association for Computational Linguistics (Volume 1: Long Papers)}, pp.\  889--898, Melbourne, Australia, 2018. Association for Computational Linguistics.
\newblock \doi{10.18653/v1/P18-1082}.
\newblock URL \url{https://aclanthology.org/P18-1082}.

\bibitem[Fan et~al.(2023)Fan, Hua, Li, Ling, Zhang, and Hemphill]{fan2023nphardeval}
Lizhou Fan, Wenyue Hua, Lingyao Li, Haoyang Ling, Yongfeng Zhang, and Libby Hemphill.
\newblock Nphardeval: Dynamic benchmark on reasoning ability of large language models via complexity classes.
\newblock \emph{arXiv preprint arXiv:2312.14890}, 2023.

\bibitem[Fang et~al.(2024)Fang, Gao, Ren, Chen, Verberne, and Ren]{fang2024multi}
Jiabao Fang, Shen Gao, Pengjie Ren, Xiuying Chen, Suzan Verberne, and Zhaochun Ren.
\newblock A multi-agent conversational recommender system.
\newblock \emph{arXiv preprint arXiv:2402.01135}, 2024.

\bibitem[Google(2024)]{gemma2}
Google.
\newblock Gemma 2: Improving open language models at a practical size, 2024.
\newblock URL \url{https://arxiv.org/abs/2408.00118}.

\bibitem[Gusev(2024)]{gusev2024pingpong}
Ilya Gusev.
\newblock Pingpong: A benchmark for role-playing language models with user emulation and multi-model evaluation.
\newblock \emph{arXiv preprint arXiv:2409.06820}, 2024.

\bibitem[He et~al.(2024{\natexlab{a}})He, Pan, Dong, Song, LiuYiBo, Qianguosun, Liang, Wang, Zhang, and Zhang]{he-etal-2024-never}
Junqing He, Kunhao Pan, Xiaoqun Dong, Zhuoyang Song, LiuYiBo LiuYiBo, Qianguosun Qianguosun, Yuxin Liang, Hao Wang, Enming Zhang, and Jiaxing Zhang.
\newblock Never lost in the middle: Mastering long-context question answering with position-agnostic decompositional training.
\newblock In Lun-Wei Ku, Andre Martins, and Vivek Srikumar (eds.), \emph{Proceedings of the 62nd Annual Meeting of the Association for Computational Linguistics (Volume 1: Long Papers)}, pp.\  13628--13642, Bangkok, Thailand, August 2024{\natexlab{a}}. Association for Computational Linguistics.
\newblock \doi{10.18653/v1/2024.acl-long.736}.
\newblock URL \url{https://aclanthology.org/2024.acl-long.736}.

\bibitem[He et~al.(2024{\natexlab{b}})He, Zeng, He, Liang, and Xiao]{he2024complex}
Qianyu He, Jie Zeng, Qianxi He, Jiaqing Liang, and Yanghua Xiao.
\newblock From complex to simple: Enhancing multi-constraint complex instruction following ability of large language models.
\newblock \emph{arXiv preprint arXiv:2404.15846}, 2024{\natexlab{b}}.

\bibitem[Hendrycks et~al.()Hendrycks, Burns, Basart, Zou, Mazeika, Song, and Steinhardt]{hendrycksmeasuring}
Dan Hendrycks, Collin Burns, Steven Basart, Andy Zou, Mantas Mazeika, Dawn Song, and Jacob Steinhardt.
\newblock Measuring massive multitask language understanding.
\newblock In \emph{International Conference on Learning Representations}.

\bibitem[Jiang et~al.(2024)Jiang, Sablayrolles, Roux, Mensch, Savary, Bamford, Chaplot, Casas, Hanna, Bressand, et~al.]{jiang2024mixtralexperts}
Albert~Q Jiang, Alexandre Sablayrolles, Antoine Roux, Arthur Mensch, Blanche Savary, Chris Bamford, Devendra~Singh Chaplot, Diego de~las Casas, Emma~Bou Hanna, Florian Bressand, et~al.
\newblock Mixtral of experts.
\newblock \emph{arXiv preprint arXiv:2401.04088}, 2024.

\bibitem[Koco{\'n} et~al.(2023)Koco{\'n}, Cichecki, Kaszyca, Kochanek, Szydlo, Baran, Bielaniewicz, Gruza, Janz, Kanclerz, et~al.]{kocon2023chatgpt}
Jan Koco{\'n}, Igor Cichecki, Oliwier Kaszyca, Mateusz Kochanek, Dominika Szydlo, Joanna Baran, Julita Bielaniewicz, Marcin Gruza, Arkadiusz Janz, Kamil Kanclerz, et~al.
\newblock Chatgpt: Jack of all trades, master of none.
\newblock \emph{Information Fusion}, 99:\penalty0 101861, 2023.

\bibitem[Li et~al.(2023)Li, Yan, Wang, Tang, Ren, Srinivasan, and Jin]{instruction-follow-verbalizer}
Shiyang Li, Jun Yan, Hai Wang, Zheng Tang, Xiang Ren, Vijay Srinivasan, and Hongxia Jin.
\newblock Instruction-following evaluation through verbalizer manipulation, 2023.

\bibitem[Li et~al.(2024)Li, Li, Cui, Bi, Wang, Wang, Yang, Shi, and Zhang]{mage}
Yafu Li, Qintong Li, Leyang Cui, Wei Bi, Zhilin Wang, Longyue Wang, Linyi Yang, Shuming Shi, and Yue Zhang.
\newblock {MAGE}: Machine-generated text detection in the wild.
\newblock In Lun-Wei Ku, Andre Martins, and Vivek Srikumar (eds.), \emph{Proceedings of the 62nd Annual Meeting of the Association for Computational Linguistics (Volume 1: Long Papers)}, pp.\  36--53, Bangkok, Thailand, August 2024. Association for Computational Linguistics.
\newblock \doi{10.18653/v1/2024.acl-long.3}.
\newblock URL \url{https://aclanthology.org/2024.acl-long.3}.

\bibitem[Meta(2024)]{llama3}
Meta.
\newblock The llama 3 herd of models, 2024.
\newblock URL \url{https://arxiv.org/abs/2407.21783}.

\bibitem[Mostafazadeh et~al.(2016)Mostafazadeh, Chambers, He, Parikh, Batra, Vanderwende, Kohli, and Allen]{roc}
Nasrin Mostafazadeh, Nathanael Chambers, Xiaodong He, Devi Parikh, Dhruv Batra, Lucy Vanderwende, Pushmeet Kohli, and James Allen.
\newblock A corpus and cloze evaluation for deeper understanding of commonsense stories.
\newblock In \emph{Proceedings of the 2016 Conference of the North {A}merican Chapter of the Association for Computational Linguistics: Human Language Technologies}, pp.\  839--849, San Diego, California, 2016. Association for Computational Linguistics.
\newblock \doi{10.18653/v1/N16-1098}.
\newblock URL \url{https://aclanthology.org/N16-1098}.

\bibitem[Narayan et~al.(2018)Narayan, Cohen, and Lapata]{xsum}
Shashi Narayan, Shay~B Cohen, and Mirella Lapata.
\newblock Don't give me the details, just the summary! topic-aware convolutional neural networks for extreme summarization.
\newblock \emph{arXiv preprint arXiv:1808.08745}, 2018.

\bibitem[Ouyang et~al.(2022)Ouyang, Wu, Jiang, Almeida, Wainwright, Mishkin, Zhang, Agarwal, Slama, Ray, Schulman, Hilton, Kelton, Miller, Simens, Askell, Welinder, Christiano, Leike, and Lowe]{instruct-gpt}
Long Ouyang, Jeff Wu, Xu~Jiang, Diogo Almeida, Carroll~L. Wainwright, Pamela Mishkin, Chong Zhang, Sandhini Agarwal, Katarina Slama, Alex Ray, John Schulman, Jacob Hilton, Fraser Kelton, Luke Miller, Maddie Simens, Amanda Askell, Peter Welinder, Paul Christiano, Jan Leike, and Ryan Lowe.
\newblock Training language models to follow instructions with human feedback, 2022.

\bibitem[Park et~al.(2024)Park, Liu, Ozdaglar, and Zhang]{park2024llm}
Chanwoo Park, Xiangyu Liu, Asuman~E Ozdaglar, and Kaiqing Zhang.
\newblock Do llm agents have regret? a case study in online learning and games.
\newblock In \emph{ICLR 2024 Workshop on Large Language Model (LLM) Agents}, 2024.

\bibitem[{Qwen Team}(2024)]{qwen2.5}
{Qwen Team}.
\newblock Qwen2.5: A party of foundation models, September 2024.
\newblock URL \url{https://qwenlm.github.io/blog/qwen2.5/}.

\bibitem[Rajpurkar et~al.(2016)Rajpurkar, Zhang, Lopyrev, and Liang]{squad}
Pranav Rajpurkar, Jian Zhang, Konstantin Lopyrev, and Percy Liang.
\newblock {SQ}u{AD}: 100,000+ questions for machine comprehension of text.
\newblock In \emph{Proceedings of the 2016 Conference on Empirical Methods in Natural Language Processing}, pp.\  2383--2392, Austin, Texas, 2016. Association for Computational Linguistics.
\newblock \doi{10.18653/v1/D16-1264}.
\newblock URL \url{https://aclanthology.org/D16-1264}.

\bibitem[Skopek et~al.(2023)Skopek, Aralikatte, Gooding, and Carbune]{instruction-follow-summ}
Ondrej Skopek, Rahul Aralikatte, Sian Gooding, and Victor Carbune.
\newblock Towards better evaluation of instruction-following: A case-study in summarization.
\newblock In Jing Jiang, David Reitter, and Shumin Deng (eds.), \emph{Proceedings of the 27th Conference on Computational Natural Language Learning (CoNLL)}, pp.\  221--237, Singapore, December 2023. Association for Computational Linguistics.
\newblock \doi{10.18653/v1/2023.conll-1.16}.
\newblock URL \url{https://aclanthology.org/2023.conll-1.16}.

\bibitem[Suzgun et~al.(2023)Suzgun, Scales, Sch{\"a}rli, Gehrmann, Tay, Chung, Chowdhery, Le, Chi, Zhou, et~al.]{suzgun2023challenging}
Mirac Suzgun, Nathan Scales, Nathanael Sch{\"a}rli, Sebastian Gehrmann, Yi~Tay, Hyung~Won Chung, Aakanksha Chowdhery, Quoc Le, Ed~Chi, Denny Zhou, et~al.
\newblock Challenging big-bench tasks and whether chain-of-thought can solve them.
\newblock In \emph{Findings of the Association for Computational Linguistics: ACL 2023}, pp.\  13003--13051, 2023.

\bibitem[Tan et~al.(2016)Tan, Niculae, Danescu{-}Niculescu{-}Mizil, and Lee]{cmv}
Chenhao Tan, Vlad Niculae, Cristian Danescu{-}Niculescu{-}Mizil, and Lillian Lee.
\newblock Winning arguments: Interaction dynamics and persuasion strategies in good-faith online discussions.
\newblock In Jacqueline Bourdeau, Jim Hendler, Roger Nkambou, Ian Horrocks, and Ben~Y. Zhao (eds.), \emph{Proceedings of the 25th International Conference on World Wide Web, {WWW} 2016, Montreal, Canada, April 11 - 15, 2016}, pp.\  613--624. {ACM}, 2016.
\newblock \doi{10.1145/2872427.2883081}.
\newblock URL \url{https://doi.org/10.1145/2872427.2883081}.

\bibitem[Taori et~al.(2023)Taori, Gulrajani, Zhang, Dubois, Li, Guestrin, Liang, and Hashimoto]{alpaca}
Rohan Taori, Ishaan Gulrajani, Tianyi Zhang, Yann Dubois, Xuechen Li, Carlos Guestrin, Percy Liang, and Tatsunori~B. Hashimoto.
\newblock Stanford alpaca: An instruction-following llama model.
\newblock \url{https://github.com/tatsu-lab/stanford_alpaca}, 2023.

\bibitem[Touvron et~al.(2023)Touvron, Martin, Stone, Albert, Almahairi, Babaei, Bashlykov, Batra, Bhargava, Bhosale, Bikel, Blecher, Ferrer, Chen, Cucurull, Esiobu, Fernandes, Fu, Fu, Fuller, Gao, Goswami, Goyal, Hartshorn, Hosseini, Hou, Inan, Kardas, Kerkez, Khabsa, Kloumann, Korenev, Koura, Lachaux, Lavril, Lee, Liskovich, Lu, Mao, Martinet, Mihaylov, Mishra, Molybog, Nie, Poulton, Reizenstein, Rungta, Saladi, Schelten, Silva, Smith, Subramanian, Tan, Tang, Taylor, Williams, Kuan, Xu, Yan, Zarov, Zhang, Fan, Kambadur, Narang, Rodriguez, Stojnic, Edunov, and Scialom]{touvron2023llama}
Hugo Touvron, Louis Martin, Kevin Stone, Peter Albert, Amjad Almahairi, Yasmine Babaei, Nikolay Bashlykov, Soumya Batra, Prajjwal Bhargava, Shruti Bhosale, Dan Bikel, Lukas Blecher, Cristian~Canton Ferrer, Moya Chen, Guillem Cucurull, David Esiobu, Jude Fernandes, Jeremy Fu, Wenyin Fu, Brian Fuller, Cynthia Gao, Vedanuj Goswami, Naman Goyal, Anthony Hartshorn, Saghar Hosseini, Rui Hou, Hakan Inan, Marcin Kardas, Viktor Kerkez, Madian Khabsa, Isabel Kloumann, Artem Korenev, Punit~Singh Koura, Marie-Anne Lachaux, Thibaut Lavril, Jenya Lee, Diana Liskovich, Yinghai Lu, Yuning Mao, Xavier Martinet, Todor Mihaylov, Pushkar Mishra, Igor Molybog, Yixin Nie, Andrew Poulton, Jeremy Reizenstein, Rashi Rungta, Kalyan Saladi, Alan Schelten, Ruan Silva, Eric~Michael Smith, Ranjan Subramanian, Xiaoqing~Ellen Tan, Binh Tang, Ross Taylor, Adina Williams, Jian~Xiang Kuan, Puxin Xu, Zheng Yan, Iliyan Zarov, Yuchen Zhang, Angela Fan, Melanie Kambadur, Sharan Narang, Aurelien Rodriguez, Robert Stojnic, Sergey Edunov, and Thomas
  Scialom.
\newblock Llama 2: Open foundation and fine-tuned chat models, 2023.

\bibitem[Wang et~al.()Wang, Tang, Zhao, Wang, and Wen]{wang2023rethinking}
Xiaolei Wang, Xinyu Tang, Xin Zhao, Jingyuan Wang, and Ji-Rong Wen.
\newblock Rethinking the evaluation for conversational recommendation in the era of large language models.
\newblock In \emph{The 2023 Conference on Empirical Methods in Natural Language Processing}.

\bibitem[Xiao et~al.()Xiao, Tian, Chen, Han, and Lewis]{xiaoefficient}
Guangxuan Xiao, Yuandong Tian, Beidi Chen, Song Han, and Mike Lewis.
\newblock Efficient streaming language models with attention sinks.
\newblock In \emph{The Twelfth International Conference on Learning Representations}.

\bibitem[Yan et~al.(2024)Yan, Luo, and Zhang]{yan2024refutebench}
Jianhao Yan, Yun Luo, and Yue Zhang.
\newblock Refutebench: Evaluating refuting instruction-following for large language models.
\newblock \emph{arXiv preprint arXiv:2402.13463}, 2024.

\bibitem[Zellers et~al.(2019)Zellers, Holtzman, Bisk, Farhadi, and Choi]{hswag}
Rowan Zellers, Ari Holtzman, Yonatan Bisk, Ali Farhadi, and Yejin Choi.
\newblock {H}ella{S}wag: Can a machine really finish your sentence?
\newblock In \emph{Proceedings of the 57th Annual Meeting of the Association for Computational Linguistics}, pp.\  4791--4800, Florence, Italy, 2019. Association for Computational Linguistics.
\newblock \doi{10.18653/v1/P19-1472}.
\newblock URL \url{https://aclanthology.org/P19-1472}.

\bibitem[Zheng et~al.(2023)Zheng, Chiang, Sheng, Zhuang, Wu, Zhuang, Lin, Li, Li, Xing, Zhang, Gonzalez, and Stoica]{mt-bench}
Lianmin Zheng, Wei-Lin Chiang, Ying Sheng, Siyuan Zhuang, Zhanghao Wu, Yonghao Zhuang, Zi~Lin, Zhuohan Li, Dacheng Li, Eric~P. Xing, Hao Zhang, Joseph~E. Gonzalez, and Ion Stoica.
\newblock Judging llm-as-a-judge with mt-bench and chatbot arena, 2023.
\newblock URL \url{https://arxiv.org/abs/2306.05685}.

\bibitem[Zhou et~al.(2023)Zhou, Lu, Mishra, Brahma, Basu, Luan, Zhou, and Hou]{zhou2023instruction}
Jeffrey Zhou, Tianjian Lu, Swaroop Mishra, Siddhartha Brahma, Sujoy Basu, Yi~Luan, Denny Zhou, and Le~Hou.
\newblock Instruction-following evaluation for large language models.
\newblock \emph{arXiv preprint arXiv:2311.07911}, 2023.

\bibitem[Zhu et~al.(2023)Zhu, Chen, Wang, Gong, Yang, and Xie]{zhu2023dyval}
Kaijie Zhu, Jiaao Chen, Jindong Wang, Neil~Zhenqiang Gong, Diyi Yang, and Xing Xie.
\newblock Dyval: Graph-informed dynamic evaluation of large language models.
\newblock \emph{arXiv preprint arXiv:2309.17167}, 2023.

\bibitem[Zhu et~al.(2024)Zhu, Wang, Zhao, Xu, and Xie]{zhu2024dyval}
Kaijie Zhu, Jindong Wang, Qinlin Zhao, Ruochen Xu, and Xing Xie.
\newblock Dyval 2: Dynamic evaluation of large language models by meta probing agents.
\newblock \emph{arXiv preprint arXiv:2402.14865}, 2024.

\end{thebibliography}
\bibliographystyle{iclr2025_conference}


\end{document}